# Cost-Aware Query Policies in Active Learning for Efficient Autonomous Robotic Exploration*

Sapphira Akins, Hans Mertens, Frances Zhu, *Member, IEEE*

*Abstract*— In missions constrained by finite resources efficient data collection is critical. Informative path planning, driven by automated decision-making, optimizes exploration by reducing the costs associated with accurate characterization of a target in an environment. Previous implementations of active learning (AL) did not consider the action cost for regression problems or only considered the action cost for classification problems. This paper analyzes an AL algorithm for Gaussian Process (GP) regression while incorporating action cost. The algorithm's performance is compared on various regression problems to include terrain mapping on diverse simulated surfaces along metrics of root mean square (RMS) error, samples and distance until convergence, and model variance upon convergence. The cost-dependent acquisition policy doesn't organically optimize information gain over distance; instead, the traditional uncertainty metric with a distance constraint best minimizes root-mean-square error over trajectory distance. This study's impact is to provide insight into incorporating action cost with AL methods to optimize exploration under realistic mission constraints.

## I. INTRODUCTION

In the field of machine learning and robotics, efficiently gathering data is crucial for driving technological advancements. As the demand for data to train machine learning models grows, so do the costs associated with collecting and processing that data. To optimize these efforts, we must move away from indiscriminate data acquisition and focus on strategies that minimize costs while maximizing information gain. Active learning offers a solution by intelligently selecting the most informative data points, reducing the need for excessive sampling and making data collection more cost-effective. The benefits of active learning include improved prediction accuracy, faster convergence to a learned model, reduced data acquisition costs, improved generalization, and robustness to noise [5], [6].

This challenge is especially critical in a multitude of diverse applications, such as ocean exploration, where only 5-20% of the oceans have been explored due to the harsh and inaccessible environments [1]. Similarly, space missions are expensive and resource-constrained, requiring careful selection of exploration targets to maximize scientific return within limited mission lifetimes [2]. In agriculture, the need to monitor vast swaths of land, often beyond the reach of remote sensing technologies, calls for efficient, targeted sampling strategies [3]. Even in domains like material discovery, where new materials must be identified from a vast search space and prototyping new materials is costly, intelligently selecting experiments can drastically reduce costs and speed up breakthroughs [4]. Across these fields, the ability to sample data intelligently is not only a matter of efficiency but also a necessity for advancing technology within practical constraints.

Active learning holds immense promise in maximizing information gain, but much progress in the field of active learning ignores the cost of annotating unlabeled data or measuring a sample [7]. The act of annotation occurs when a proposed data point in the input space is presented to an annotator to label; for example, an image to a human to classify as an object class or a location of interest to a robot's sensor to measure a real value. Labeling these samples consumes human, computational, or robotic effort, which should be incorporated into the active learning framework. Applications in which annotation cost is highly emphasized include exploratory robotics, such as extraplanetary rovers mapping the Moon [2], [8] and underwater autonomous vehicles mapping the ocean [9], [10]. Practical and scalable active learning algorithms for robotic applications and beyond must include annotation costs to be implemented in reality.

This paper aims to answer the following questions: i) Does considering annotation cost balance information gain and annotation cost? What is the performance gain? ii) What is the annotation-cost query policy that best balances cost and information gain? This paper's contribution is showing that an annotation-conscious query policy does indeed balance cost and information gain, while guiding practitioners on how to design a query policy to incorporate annotation costs to achieve balance.

Section 2 below provides further background into active learning methods and their characterization. Section 3 follows with the methodology which outlines the experiment design, procedures, and campaign, the model hyperparameters, and the benchmark testing surfaces. Section 4 discusses the evaluation metrics and Section 5 presents data gathered from two experiments conducted in virtual environments. Finally, Section 6 summarizes the results and provides recommendations for future courses of action.

## II. BACKGROUND

Active learning is a well-established framework aimed at optimizing data acquisition by selecting the most informative data points with minimal labeling or annotation costs. While

*Research supported by the National Science Foundation (NSF).

S. Akins is with the University of Hawai'i at Manoa, Honolulu, HI 96822 USA (e-mail: sakins@hawaii.edu).

H. Mertens is with the University of Hawai'i at Manoa, Honolulu, HI 96822 USA (e-mail: hansm@hawaii.edu).

F. Zhu is with the University of Hawai'i at Manoa, Honolulu, HI 96822 USA (e-mail: zhuf@hawaii.edu).

this approach is valuable in domains where human annotation is costly, it does not fully address the challenges posed by physical exploration tasks where action costs, such as movement or sampling, must also be considered. In the context of spatial exploration, Gaussian Processes (GPs) have been shown to be highly effective in active learning frameworks for optimizing data acquisition. Krause and Guestrin introduce a nonmyopic active learning strategy using GPs, focusing on the trade-off between exploration (uncertainty reduction) and exploitation (near-optimal observation selection) [11], [12]. Their approach demonstrates how GPs can be used to efficiently select observations in spatial domains, which is directly relevant to exploration tasks such as planetary missions. While their work optimizes the exploration-exploitation trade-off, it does not account for action cost. Action costs refer to the physical movement of an autonomous system, which is a key consideration in planetary exploration. This gap highlights the need for further integration of action costs into the decision-making framework for autonomous exploration.

Active learning (AL) is an iterative process designed to optimize data acquisition by selecting the next sample point in an input space based on a query policy. This policy aims to maximize an objective, such as reducing uncertainty or minimizing prediction error. AL is composed of several key components: the learner, annotator, and query policy. The learner processes historical data to make predictions, while the annotator provides labels or ground truth for selected samples. The query policy determines which points should be labeled next, optimizing data acquisition based on an information criterion. In most conventional AL frameworks, the next sampling point $x_{i+1}$ is chosen by maximizing a query policy $g(\hat{f})$ as seen in Eq. (1).

$$x_{i+1} = \underset{x \in D_U}{\mathrm{argmax}}\, g(\hat{f}(x)) \qquad (1)$$

$$\hat{f}(x) \sim \mathcal{N}(m(x), k(x,x)) \qquad (2)$$

The learner $\hat{f}(x)$ is defined by Eq. (2). $m(x)$ and $k(x,x)$ represent the mean and covariance functions, respectively. $D_U$ is the unlabeled data space. $g(x)$ is the acquisition function, conventionally in the form of information criteria.

The learner becomes more accurate as it receives more data, while the annotator incurs costs for each labeled point. Annotation costs can be uniform, where labeling a point causes the same cost regardless of spatial location (spatially independent). However, annotation costs can be variable, influenced by factors like the distance between sampled points. This is crucial in fields like planetary exploration, where moving between locations has a tangible cost.

The query policy, critical in determining the balance between exploration (sampling from uncertain areas) and exploitation (focusing on well-understood regions), often prioritizes information gain. However, in physically constrained environments, solely maximizing information gain is impractical. Such applications require factoring in both data informativeness and action costs, such as movement or energy usage, when deciding where to sample next. Without accounting for action costs, the algorithm may select points that are informative but too costly to reach, reducing overall mission efficiency.

GP AL algorithms excel in scenarios involving sparse and unevenly distributed data; however, they can struggle with larger datasets due to computational demands. Despite this challenge, GPs have been shown to outperform other models, such as Bayesian Neural Networks (BNNs) in sparse or low dimensional data regimes or in expressing smooth target manifolds [8]. This paper builds on the work done in [8], [2] studying active learning for planetary exploration driven by an information query policy. This study focuses on the optimization of the query policy. An ideal algorithm should not only consider sampling efficiency, but also the cost of the respective samples. By integrating both uncertainty and distance metrics into the acquisition function, the algorithm accurately balances the 'cost' and 'reward' of its sampling actions. This set-up allows for an AL algorithm to choose actions that optimize the trade-off between minimizing distance traveled and maximizing information gain.

### III. METHODOLOGY

The following sections address the query policies tested, the AL algorithm and the GP model's hyperparameters. Subsequently, the benchmark surfaces and the experimental campaign are presented.

#### A. Query Policies

In this study, the agent traversing the surface is encoded with an objective function that aims to minimize a learned model's prediction with respect to the ground truth. The model error takes on the form of the $\mathcal{L}_2$ norm, also known as root mean squared (RMS) error. The query policy or acquisition function, which guides the agent's sampling decision, can be configured to send the rover to sampling locations based on various criteria. This study compares a traditional query policy configuration (i.e., sampling at the location of highest variance regardless of distance from the current location) and novel policies that weigh both variance and distance required to reach a sampling location: distance-normalized variance and distance constrained variance policies.

A conventional uncertainty sampling policy can be expressed by Eq. (3)

$$x_{i+1} = \underset{x_j \in D_U}{\mathrm{argmax}}\, \sigma^2_{pred}(x_j) \qquad (3)$$

$$\sigma^2_{pred}(x) = diag(k(x,x)) \qquad (4)$$

where $\sigma^2_{pred}(x)$ represents the model variance across the unlabeled data set $D_U$.

The distance-constrained variance query policy determines the next sampling location, which can be restricted to a certain movement horizon set by the algorithm. The concept of a movement horizon is particularly important when considering different types of technology. For example, a satellite would use an unconstrained movement horizon as it can point to any location within a space with relatively low cost. In contrast, a rover operating in-situ will have much higher costs associated with moving from point to point. In this case, the movement horizon can be constrained to a certain grouping of nearby points. The distance-normalized policy and the distance-constrained policy are outlined in Eq. (5) and Eq. (6), respectively.

$$x_{i+1} = \underset{x_j \in D_u}{\operatorname{argmax}} \sigma^2_{pred}(x_j)/\|x_j - x_i\| \quad (5)$$

$$x_{i+1} = \underset{\|x_j - x_i\| \leq r_{con}}{\operatorname{argmax}} \sigma^2_{pred}(x_j) \quad (6)$$

Various distance constraints, or movement horizons, were tested and compared on their effectiveness of decreasing distance traveled and samples necessary for convergence to a low RMSE model of the sampled space for the distance-constrained policy. Specifically, these distance constraints include: $1\Delta x, 2\Delta x, 3\Delta x, 5\Delta x, 7\Delta x$, and $10\Delta x$.

The distance-normalized variance policy uses distance-normalized variance in its query policy. The agent's movement while operating under this exploration strategy is unconstrained, as the model is expected to determine the most cost-effective and rewarding point to travel to within the entire sample space.

### B. Algorithm

The active learning algorithm utilized for both distance-normalized and distance-constrained variance query policies is outlined in Algorithm 1. Note that all experiments follow the same steps, except for any deviations which are mentioned explicitly. All variables mentioned in Algorithm 1 are defined in Table 1.

---

**Algorithm 1** Gaussian Process Active Learning

1: Define environemnt $\in \{Parabola, Townsend, Lunar\}$
2: Define spatial limits $(x_{min}, x_{max}, y_{min}, y_{max})$
3: Define query policy $\in \{dist_{constrained}, dist_{normalized}, conventional\}$
4: Define stopping condition $gridpoints_{total}/4$
5: Define noise $\sigma^2_{noise}$
6: Define movement horizon
7: Define target output distribution
8: Initialize position in environment
9: **for** $i = 1$ to 10
10:    Randomly choose $x_i \in D_U$ constrained to $3\Delta x$
11:    Measure ground truth $y$ at $x_{i+1}$
12:    Add $(x_i, y_i)$ to $D_{train}$
13: Train GP model on $D_{train}$
14: **for** $i = 11$ to $n_{samples}$
15:    Train GP model on $D_{train}$
16:    Predict $\hat{Y}_{pred}$ and variance $\sigma^2_{pred}$ in prediction horizon $r_{con}$
17:    Use query policy $g^*$ to find $x_{i+1}$
18:    Traverse to $x_{i+1}$ to measure ground truth $y$
19:    Add $(x_{i+1}, y_{i+1})$ to $D_{train}$

---

TABLE I.    DEFINITION OF VARIABLES IN ALGORITHM 1

| Variable | Description |
|---|---|
| $DEM_{resolution}$ | Digital Elevation Map (DEM) resolution, defined as 5 meters for grid sizing and 250 meters for hydroxyl content |
| $D_U$ | Unsampled data points across surface geometry |
| $x_{start}$ | The starting position of the agent on the surface |
| $D_{train}$ | Training data set consisting of a sampling location and the ground truth observation |
| $\hat{Y}_{pred}$ | Predicted scalar expected values |
| $\sigma^2_{pred}$ | Predicted variance values |
| $r_{con}$ | Movement horizon |

### C. Model Hyperparameters

The AL Algorithm was developed using the GPyTorch package [4]. During simulations, the GP model uses a Radial Basis Function (RBF) kernel. The kernel's length scale is optimized through gradient descent over 100 iterations, with the model trained to maximize the marginal log likelihood. The model's hyperparameters were optimized using the Adam optimizer with a learning rate of 0.1. The code utilized in this study is available in the following repository: https://github.com/xfyna/Action_Cost_AL.git. All computations were performed on KOA, the University of Hawaii's high-performance computing (HPC) cluster.

### D. Experiment Environments

The exploration strategies are evaluated through their capabilities of learning three spatial distributions of a target output of varying complexity by traversing a geometric surface shown in Fig. 1 and 2 below. These include the Parabola, Townsend, and Lunar Crater surface geometries. The changing complexity of these surfaces allows the exploration strategies to be evaluated on their convergence rates and various other evaluation metrics, explained in more detail in the following section. Note that Fig. 1 represents the surface the agent traverses and therefore is used to calculate the distance the agent travels. Fig. 2, on the other hand, is the target output distribution that model aims to learn. In the case of the Parabola and Townsend surfaces, the agent is building a model that learns the elevation of the surface. In the case of the Lunar surface, the model learns the hydroxyl content across the surface.

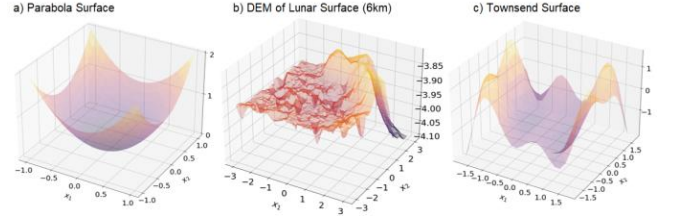

Figure 1. Surface environments that the agent traverses which dictates distance traveled by the agent: a) Parabola, b) Lunar Crater (6 km edge crater elevation), c) Townsend.

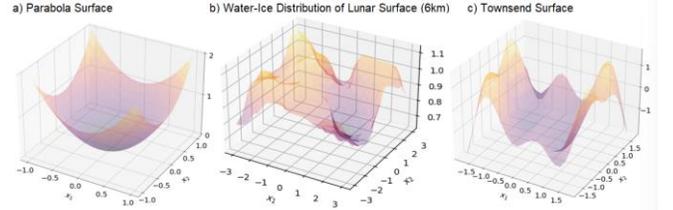

Figure 2. True value of the target outputs the agent learns that dictates RMS error: a) Parabola elevation, b) LAMP data across the DEM for the 6 km crater swath, c) Townsend elevation.

The surfaces are characterized by two independent dimensions (planar position $r = (x_1, x_2)$) and a third dependent dimension $y$. For the Parabola and Townsend surfaces, the algebraic relationship between $y$ and position is simulated. For the Lunar Crater, this target is taken from real measurements, with $y$ representing the hydroxyl content across the surface rather than "elevation", as it does for the previous two surfaces. The Parabola surface is described by Eq. (7), where $x_1$ and $x_2$ range from $[-1: 0.1: 1]$. The

Townsend surface is defined by Eq. (8), with $x_1$ and $x_2$ ranging from $[-2.5: 0.1: 2.5]$.

$$y = x_1^2 + x_2^2 + \sigma_{noise}^2 \quad (7)$$

$$y = -(\cos((x_1 - 0.1)x_2) - x_1 \sin(3x_1 + x_2) + \sigma_{noise}^2 \quad (8)$$

The Lunar Crater surface is derived from the Lyman Alpha Mapping Project (LAMP) data which collected ultraviolet spectrometry data from the Lunar Reconnaissance Orbiter (LRO) [5]. The dataset includes a digital elevation map (DEM) of the lunar south pole, represented by $r = (x_1, x_2, x_3)$, with a spatial resolution of 5 m and hydroxyl data with a resolution of 250 m. The data contains varying levels of noise, and substantial gaps are evident near the crater rim.

The complexity of these surfaces is defined by the number of local wells and measurement noise, seen in Table 2 in ascending complexity. The code used to calculate the number of wells can be found in the previously linked GitHub repository.

TABLE II. RANGE OF SIMULATED ENVIRONMENT

| Environment Topography | Size of Surface | # of Local Extrema | Known Expression |
|---|---|---|---|
| Noisy Parabola | $x_1 \in [-1: 0.1: 1]$ | Min: 1 | Yes |
| | $x_2 \in [-1: 0.1: 1]$ | Max: 0 | |
| Noisy Lunar Crater | $x_1 \in [-3: 0.25: 3]$ | Min: 5 | Yes |
| | $x_2 \in [-3: 0.25: 3]$ | Max: 3 | |
| Noisy Townsend | $x_1 \in [-2.5: 0.25: 2.5]$ | Min: 6 | No |
| | $x_2 \in [-2.5: 0.25: 2.5]$ | Max: 5 | |

*E. Experimental Campaign*

A total of 10 trials were conducted for each query policy tested over each surface type. A simulation trial consists of loading the environment geometry, outlined in Table 2, followed by defining the exploration strategy, as defined in Table 3 below. The experimental simulations conducted measure the total distance the agent travels, the number of samples collected, and the model's RMSE at convergence. Note that the code is configured such that during a run, one trial is completed for all exploration strategies over the designated surface (i.e., the distance-constrained variance policy for 6 movement horizons and the distance-normalized variance policy agent traverses the same surface with random initialization across the space). As such, the surface type must be specified for each experiment, but the exploration strategies do not need to be specified when running the code.

TABLE III. RANGE OF EXPLORATION STRATEGIES AND TRIALS COLLECTED

| Exploration Strategy | Movement Horizon |
|---|---|
| Distance-Constrained Variance | $1\Delta x$ |
| | $2\Delta x$ |
| | $3\Delta x$ |
| | $5\Delta x$ |
| | $7\Delta x$ |
| | $10\Delta x$ |
| Distance-Normalized Variance | Global |
| Conventional Variance Only | Global |

IV. EVALUATION METRICS

The following single metrics are used to assess the performance of the various exploration strategies utilized in this research. RMSE upon convergence, $e_c$, is derived from control theory's concept of 2% settling time. The global RMSE between a model prediction and true values are inspected to ensure that there are enough data points for convergence to occur, as well as that the final values of RMSE stay within a 2% band of the final error, $e_f$. This 2% error band is found using Eq. (9) below. RMSE upon convergence is defined as the upper bound of this error band, as shown in Eq. (10).

$$\Delta e_{2\%} = 0.02(e_0 - e_f) \quad (9)$$

$$e_c = e_f + e_{2\%} \quad (10)$$

The normalized RMSE (NRMSE), $e_n$, is defined by the Eq. (13) where the range of target output values across the Parabola, Townsend, and Lunar surfaces are, respectively, 2.00, 5.59, 0.50.

$$e_n = \frac{e_c}{y_{max} - y_{min}} \quad (11)$$

Samples until convergence, $i_c$, denotes the index at which convergence occurs. It is calculated through the minimization of the difference between the error at an index, $i_e$, and the error upon convergence, $e_c$, as shown in Eq. (12). The values are scaled based on the number of samples taken per surface (i.e., the total number of samples taken until convergence are divided by the max possible samples per surface). The Lunar surface has a stopping condition of 155 samples and the Parabola and Townsend surface has one of 109 samples.

$$i_c = \frac{\underset{i}{\operatorname{argmin}} \|e_i - e_c\|_2}{i_{max}} \quad (12)$$

Distance until convergence, $d_c$ is a metric that represents the total distance traveled by the agent until convergence is reached. This is calculated as the sum of the radial difference between each waypoint until the sample of convergence is taken, as shown in Eq. (13). These values are scaled based on the grid length, as shown in Table 2.

$$d_c = \frac{\sum_{i=1}^{i_c} \|x_{k,i+1} - x_{k,i}\|_2}{x_{max} - x_{min}} \quad (13)$$

The following multi-objective metrics provide insight into the comparative performance of the exploration models, shedding light onto which methods perform better in terms of minimizing distance or samples while maintaining the goal of a convergence to a low error model. Distance-scaled NRMSE, $e_{dc}$, scales $e_{norm}$ by $d_c$ as shown in Eq. (14). The combined metric allows for easier interpretation regarding the various policies and their ability to converge to a low NRMSE while traveling smaller distances. The lower the value, the higher performing the model is.

$$e_{dc} = e_n * d_c \quad (14)$$

Sample-scaled NRMSE, $e_{ic}$, similarly to distance scaled NRMSE, scales the convergence NRMSE with the convergence samples, as shown in Eq. (15).

$$e_{ic} = e_n * i_c \quad (15)$$

## V. RESULTS

Considering distance in query policies can achieve similar RMS error to conventional methods but with at least a magnitude less distance traveled. Distance-constrained variance policies achieve the most distance-efficient exploration, although the most effective movement horizon depends on the environment. To illustrate these results and explore the nuances, this section presents a comprehensive metric comparison and query policy behavior.

### A. Metric Comparison

Generally, distance-constrained variance performed better than distance-normalized variance policies. Tables 4 through 6 outline the average performance metrics achieved after 10 trials with the outlined exploration methods. The green highlighted squares signify the highest scoring policy while the red-orange highlighted squares signify the lowest performing policy in the specified metric. The "Norm" movement horizon refers to the unconstrained distance-normalized variance policy and the "Conv" movement horizon denotes performance of the conventional unconstrained query policy.

TABLE IV. AVERAGE PERFORMANCE METRIC VALUES FOR ALL POLICIES ACROSS THE PARABOLA SURFACE

| Movement Horizon | $e_c$ | $i_c$ | $d_c$ | $e_{dc}$ | $e_{ic}$ |
|---|---|---|---|---|---|
| $1\Delta x$ | 0.04485 | 0.73 | 0.82750 | 0.03712 | 0.03280 |
| $2\Delta x$ | 0.01722 | 0.61 | 0.61650 | 0.01061 | 0.01042 |
| $3\Delta x$ | 0.03121 | 0.59 | 0.65200 | 0.02035 | 0.01835 |
| $5\Delta x$ | 0.02663 | 0.46 | 1.08950 | 0.02902 | 0.01227 |
| $7\Delta x$ | 0.02793 | 0.45 | 1.06300 | 0.02968 | 0.01263 |
| $10\Delta x$ | 0.02532 | 0.48 | 1.09750 | 0.02778 | 0.01215 |
| Norm | 0.00967 | 0.57 | 2.78645 | 0.02693 | 0.00546 |
| Conv | 0.00909 | 0.28 | 13.3380 | 0.12118 | 0.00253 |

TABLE V. AVERAGE PERFORMANCE METRIC VALUES FOR ALL POLICIES ACROSS THE LUNAR CRATER SURFACE

| Movement Horizon | $e_c$ | $i_c$ | $d_c$ | $e_{dc}$ | $e_{ic}$ |
|---|---|---|---|---|---|
| $1\Delta x$ | 0.10343 | 0.90 | 2.71065 | 0.28037 | 0.09350 |
| $2\Delta x$ | 0.06554 | 0.77 | 2.36771 | 0.15517 | 0.05070 |
| $3\Delta x$ | 0.05872 | 0.88 | 2.63802 | 0.15491 | 0.05171 |
| $5\Delta x$ | 0.04304 | 0.83 | 6.07708 | 0.26154 | 0.03571 |
| $7\Delta x$ | 0.04325 | 0.65 | 4.93646 | 0.21351 | 0.02827 |
| $10\Delta x$ | 0.04435 | 0.67 | 5.14236 | 0.22805 | 0.02985 |
| Norm | 0.09576 | 0.92 | 2.17940 | 0.20869 | 0.08800 |
| Conv | 0.05976 | 0.72 | 97.8479 | 5.84719 | 0.04318 |

TABLE VI. AVERAGE PERFORMANCE METRIC VALUES FOR ALL POLICIES ACROSS THE TOWNSEND SURFACE

| Movement Horizon | $e_c$ | $i_c$ | $d_c$ | $e_{dc}$ | $e_{ic}$ |
|---|---|---|---|---|---|
| $1\Delta x$ | 0.16823 | 0.92 | 2.60781 | 0.43872 | 0.15415 |
| $2\Delta x$ | 0.06573 | 0.91 | 2.69167 | 0.17692 | 0.05960 |
| $3\Delta x$ | 0.10846 | 0.86 | 2.75357 | 0.29865 | 0.09353 |
| $5\Delta x$ | 0.04700 | 0.83 | 5.34107 | 0.25105 | 0.03912 |
| $7\Delta x$ | 0.05703 | 0.81 | 5.06607 | 0.28891 | 0.04619 |
| $10\Delta x$ | 0.03989 | 0.89 | 5.69286 | 0.22707 | 0.03534 |
| Norm | 0.05611 | 0.83 | 5.68763 | 0.31914 | 0.04642 |
| Conv | 0.08521 | 0.91 | 93.4167 | 7.96004 | 0.07713 |

The best policy with respect to $e_c$ varied significantly by surface type. However, the best policies with respect to $e_{dc}$ and $e_{ic}$ were the $2\Delta x$ distance-constrained variance and $10\Delta x$ distance-constrained variance policies, respectively. Fig. 3 through 5 below highlight the balance these policies offer in terms of convergence to low error with minimal cost. The best policy with respect to $i_c$ was the $7\Delta x$ distance-constrained variance method. The best policy with respect to $d_c$ was the $2\Delta x$ distance-constrained variance method as it converged after traveling a low distance most consistently across all surfaces. On the other hand, the worst performance for both $e_c$ and $i_c$ was the $1\Delta x$ distance-constrained variance policy and the worst performance in terms of $d_c$ was the conventional unconstrained method.

Along single metrics, the $10\Delta x$, $5\Delta x$, and conventional AL policies obtained the lowest convergence NRMSE on the following surfaces, respectively: Townsend, Lunar, and Parabola. Performance in $e_c$ depends on surface complexity, with higher surface complexity resulting in higher error, as show in Fig. 3. Additionally, NRMSE depends on a policy's movement horizon where a horizon that is too small leads to increased error but once the movement horizon is sufficiently large, $e_c$ is within a couple percent of the minimum error possible. Notice that the distance-normalized and distance-constrained policies, apart from the $1\Delta x$ horizon, converge to an NRMSE that is comparable with the conventional AL method. This highlights the effectiveness of incorporating action cost into query policies, as no performance in terms of model accuracy is lost, despite the agent traveling less distance across the sampling space.

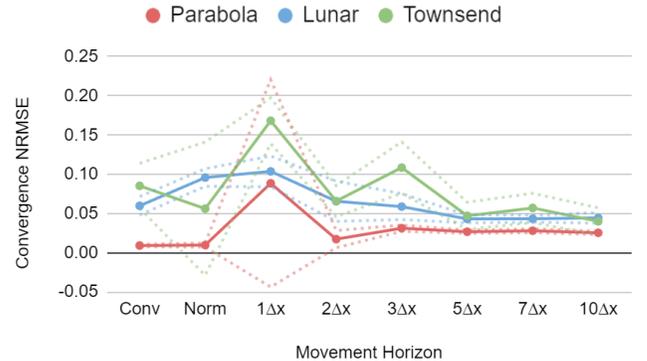

Figure 3. Average NRMSE upon Convergence across Surfaces

If only considering samples until convergence, Fig. 4 demonstrates the relationship between movement horizon and samples until convergence. The best performing metric is shown to be the $7\Delta x$ constraint, with the $10\Delta x$ trailing shortly behind. This phenomenon is observed most likely due to the exploration strategy taking larger steps allowing for exploration across a wider area, therefore not requiring as many samples to reach convergence. This is particularly evident with the conventional exploration strategy, which can be seen as taking the least samples across the parabola surface. The small dip that occurs at the $2\Delta x$ movement horizon, in Fig. 4, suggests the $2\Delta x$ constraint performs well at balancing all metrics.

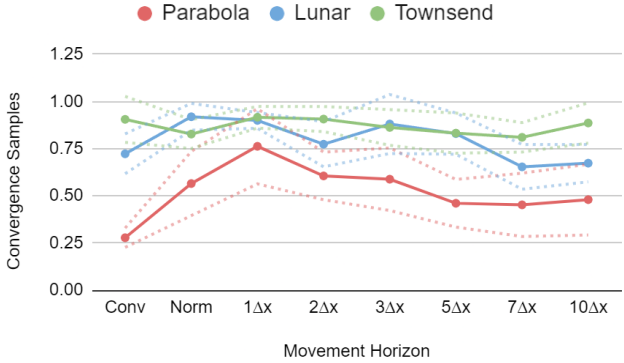

Figure 4. Average Samples Taken across Surfaces until Convergence

The best performing policies in terms of distance traveled can be seen clumped in the center of the figure at the $2\Delta x$ and $3\Delta x$ marks. Fig. 5 below illustrates the convergence distance required for the various query policies across surface types. This suggests that the movement horizon is significant in determining the distance the agent traverses until reaching convergence. Comparatively, the unconstrained strategies travel more before reaching convergence. However, the difference between the conventional unconstrained policy and the distance-normalized variance query policy is significant, shown through Fig. 5 where the conventional strategy travels an order of magnitude greater than all other methods, illustrating the importance of distance incorporation into the query policy. On average, the $5\Delta x$ constraint travels a larger distance than the other distance-constrained policies.

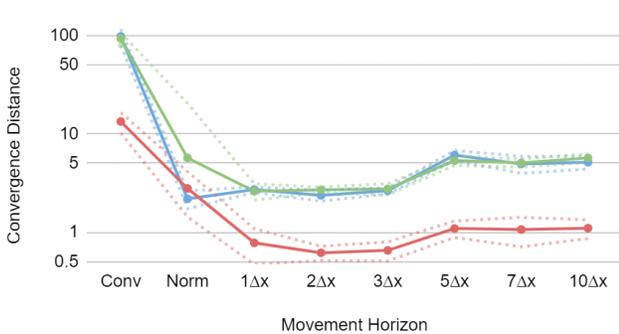

Figure 5. Average Distance Traveled across Surfaces until Convergence

To display multiple objectives, Fig. 6 and 7 present trade-offs in balace with error along the y-axis and samples, followed by distance, on the x-axis for all three surfaces. "N" represents the unconstrained distance-normalized variance method, "C" represents the conventional AL method, and all other methods are denoted by their movement constraint.

For missions that just care about error and number of samples, conventional AL makes total sense, expecially for a simple surface, but gains could be made by incorporating a movement horizon. The optimal location to balance these multiple objectives for policies would be nearest to the origin on both Figures. In Fig. 6, the Lunar surface required the highest number of samples to reach convergence for some methods irregardless of the fact that it is not the most complex surface, as seen by the blue colored labels on Fig. 6. It is possible that the kernel used to model the surface was more fitted towards the Townsend and Parabola surfaces instead of the Lunar surface causing this to occur. The effectiveness of the $7\Delta x$ constraint on limiting samples is evident as it is shown to require the least amount of samples for nearly all surface types. Along with this, its convergence NRMSE is comparable or lower than many of the other policies' convergence NRMSE. However, as mentioned previously, the $10\Delta x$ method performs the best in terms of $e_{ic}$ due to its lower convergence NRMSE and similarly low sampling rate. Looking rightward from these constraints and their locations in Fig. 6, the number of samples until convergence increases as the distance horizon decreases.

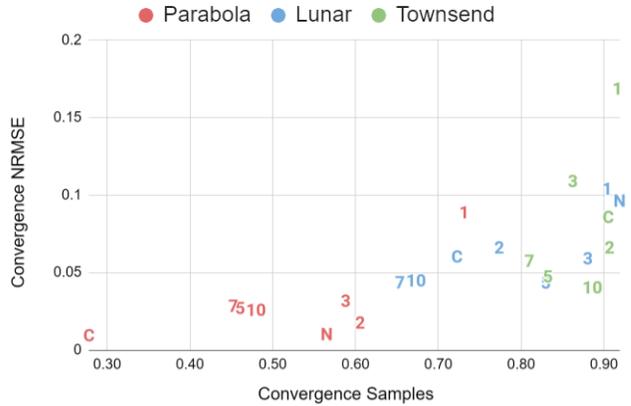

Figure 6. Mean NRMSE vs. Mean Samples across All Surfaces for 10 trials of data

Fig. 7 illustrates the clusters of policy performance in terms of distance and NRMSE across surfaces. Again, a similar visual trend can be seen with the Parabola trials hugging the leftmost side of the graph, followed by the Lunar and Townsend surface to its right (i.e., grouped by surface complexity). Further clustering is demonstrated, most visibly on the Parabola and Lunar surfaces, where lower distance horizon policies (up to $3\Delta x$) clump together at a distance metric along the horizontal axis and higher policies are shown to their bottom right, illustrating lower NRMSE convergence at higher distances. Additionally, note the separation of the conventional AL method in terms of distance traveled in comparison to NRMSE, which is generally very similar to the performance of all other models. The $2\Delta x$ method balances accuracy and samples better than all other strategies. Fig. 7 also highlights the influence of surface complexity on model performance.

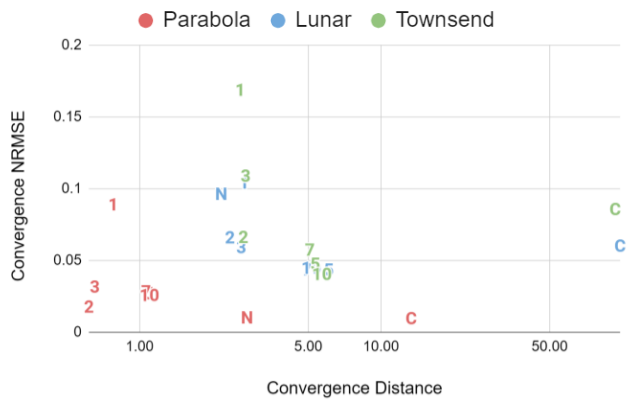

Figure 7. Mean NRMSE vs. Mean Distance until Convergence across All Surfaces for 10 trials

## B. Query Policy Behavior

During the experimental campaigns, an agent traversed a designated surface for a pre-defined number of samples. To show the variation in single trial behavior across query policies, Fig. 8 below display an example of the agent's movement across the Townsend surface using the four exploration strategies that include the distance-constrained (a & b), distance-normalized variance (c), and conventional query policies (d). The agent's path is denoted by the black line, with the star signifying its location. The semi-transparent gray surface shown over the colored surface represents the agent's internal model of the surface. While most areas of the model are shown to be nearly matched to the underlying surface, others show deficits in the agent's internal model, specifically in places where the surface is more complex.

Figure 8. Simulated Path of the Agent Traversing Townsend Surface with Various Exploration Strategies

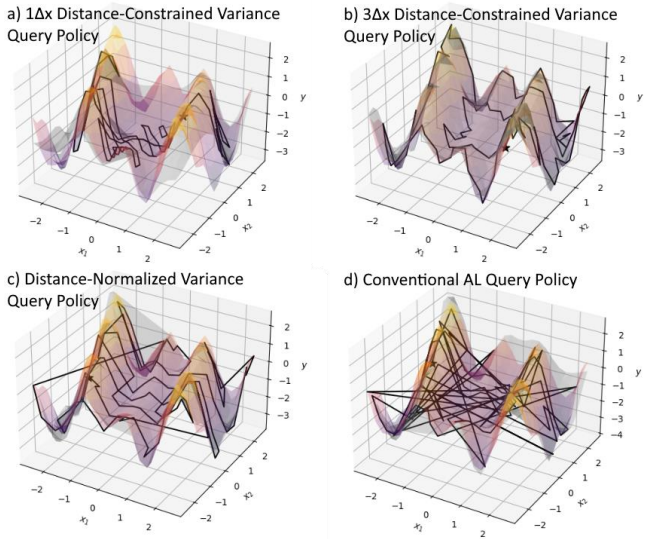

The various query policies, ordered in increasing movement horizon, show various levels of coverage and revisits across the example of the Townsend surface. The $1\Delta x$ distance-constrained variance query policy struggles to cover sufficient distance as shown in Fig. 8 a). This exploration strategy is shown as traveling in a small, clustered section of space, rather than mapping at or around the corners. The cluster location depends heavily on the initialization of the agent on the surface, leading it to be nearer to an edge or towards the center of a surface. The $3\Delta x$ distance-constrained variance query policy shown in Fig. 8 b) achieves more coverage but with a seemingly smooth path, demonstrating its distance efficiency. Moving to Fig. 8 c), the distance-normalized query policy is shown as covering more distance, however, it tends to traverse more jagged paths suggesting a lower distance efficiency. Lastly, Fig. 8 d) illustrates the conventional query policy, which is shown to move extensively across the surface revisiting many similar locations in the process.

To show the evolution of exploration, Fig. 9 displays the mean performance through a solid line and standard deviation by the shaded region across 10 Townsend trials along the metrics of both mean NRMSE and total distance traveled. Performance in terms of efficiency and accuracy can be seen through visual inspection of the area under the curve. Through this figure, the $2\Delta x$ method demonstrate its strength in traveling short distances but reducing NRMSE quickly. While the $5\Delta x$, $7\Delta x$, and $10\Delta x$ constrained policies do converge to a lower NRMSE, the distance required to reach such convergence is far greater than that for the $2\Delta x$ constraint. The limited performance of the $1\Delta x$ movement horizon is highlighted here along with the conventional AL method which is shown to perform extremely poor in comparison to all other policies in terms of convergence to a low RMS error in minimal distance.

Figure 9. Distance vs. RMS Error Mean and Stanard Deviation Variance Across 10 Trials until Convergence on the Townsend Surface

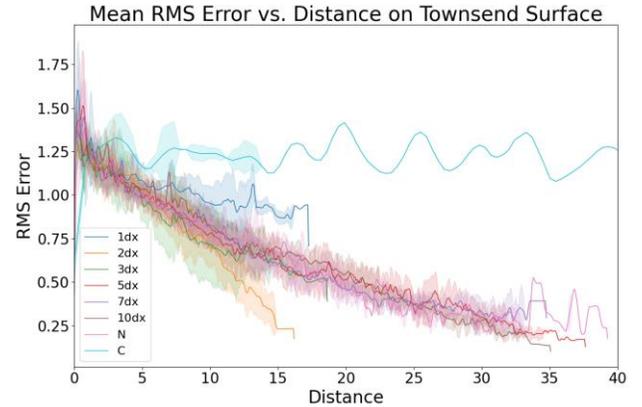

## VI. CONCLUSION

This study demonstrates the importance of incorporating action costs into active learning frameworks for autonomous robotics. By evaluating various query policies, including distance-constrained and distance-normalized variance methods, we observed that balancing information gain with movement efficiency significantly enhances mission performance. The results show that while distance-constrained policies reduce the total distance traveled without sacrificing model accuracy, the optimal movement horizon depends on the environment's complexity and mission constraints. These findings provide valuable insights for future applications, such as planetary exploration, where resource constraints and mission efficiency are paramount.

## ACKNOWLEDGMENT

This work was supported by NASA Grant HI-80NSSC21M0334 and the AI Institute in Dynamic Systems, one of the National Artificial Intelligence Research institutes funded by the National Science Foundation (NSF) via award number 2112085. We would like to thank the University of Hawaii's high-performance computing (HPC) cluster team for ensuring the operation of the HPC cluster which allowed us to complete our research, as well.

## REFERENCES

[1] B. R. C. Kennedy and R. D. Rotjan, "Mind the gap: comparing exploration effort with global biodiversity patterns and climate projections to determine ocean areas with greatest exploration needs," Front. Mar. Sci., vol. 10, Nov. 2023, doi: 10.3389/fmars.2023.1219799.
[2] A. Akemoto and F. Zhu, "Informative Path Planning to Explore and Map Unknown Planetary Surfaces with Gaussian Processes," p. 12.

[3] K. Berger, J. P. Rivera Caicedo, L. Martino, M. Wocher, T. Hank, and J. Verrelst, "A Survey of Active Learning for Quantifying Vegetation Traits from Terrestrial Earth Observation Data," Remote Sensing, vol. 13, no. 2, p. 287, Jan. 2021, doi: 10.3390/rs13020287.

[4] J. Allotey, K. T. Butler, and J. Thiyagalingam, "Entropy-based Active Learning of Graph Neural Network Surrogate Models for Materials Properties," The Journal of Chemical Physics, vol. 155, no. 17, p. 174116, Nov. 2021, doi: 10.1063/5.0065694.

[5] B. Settles, "Active Learning Literature Survey".

[6] B. Kim and B. C. Ko, "Active Contrastive Learning With Noisy Labels in Fine-Grained Classification," in 2024 International Conference on Electronics, Information, and Communication (ICEIC), Jan. 2024, pp. 1–5. doi: 10.1109/ICEIC61013.2024.10457229.

[7] B. Settles, M. Craven, and L. Friedland, "Active Learning with Real Annotation Costs".

[8] S. Akins and F. Zhu, "Comparing Active Learning Performance Driven by Gaussian Processes or Bayesian Neural Networks for Constrained Trajectory Exploration," in ASCEND 2023, in ASCEND., American Institute of Aeronautics and Astronautics, 2023. doi: 10.2514/6.2023-4720.

[9] D.-H. Cho, J.-S. Ha, S. Lee, S. Moon, and H.-L. Choi, "Informative Path Planning and Mapping with Multiple UAVs in Wind Fields," Oct. 2016.

[10] A. Anil Meera, M. Popovic, A. Millane, and R. Siegwart, Obstacle-aware Adaptive Informative Path Planning for UAV-based Target Search. 2019.

[11] A. Krause and C. Guestrin, "Nonmyopic active learning of Gaussian processes: an exploration-exploitation approach," in Proceedings of the 24th international conference on Machine learning, in ICML '07. New York, NY, USA: Association for Computing Machinery, Jun. 2007, pp. 449–456. doi: 10.1145/1273496.1273553

[12] Krause, A. Singh, and C. Guestrin, "Near-Optimal Sensor Placements in Gaussian Processes: Theory, Efficient Algorithms and Empirical Studies".